\def\year{2018}\relax
\documentclass[letterpaper]{article} 
\usepackage{aaai18}  
\usepackage{times}  
\usepackage{helvet}  
\usepackage{courier}  
\usepackage{url}  
\usepackage{graphicx}  
\usepackage{amsfonts}
\usepackage{mathtools}

\usepackage{graphicx}
\usepackage{subcaption}

\usepackage{enumitem}

\title{Selective Experience Replay for Lifelong Learning}

\author{ 
  David Isele \\
  The University of Pennsylvania, Honda Research Institute\\
  \texttt{isele@seas.upenn.edu} \\
  \And
  Akansel Cosgun \\
  Honda Research Institute \\
  \texttt{akansel.cosgun@gmail.com} \\
}
\begin{document}

\maketitle

\begin{abstract}

Deep reinforcement learning  has emerged as a powerful tool for a variety of learning tasks, however deep nets typically exhibit forgetting when learning multiple tasks in sequence. To mitigate forgetting, we propose an experience replay process that augments the standard FIFO buffer and selectively stores experiences in a long-term memory. We explore four strategies for selecting which experiences will be stored: favoring surprise, favoring reward, matching the global training distribution, and maximizing coverage of the state space. We show that distribution matching successfully prevents catastrophic forgetting, and is consistently the best approach on all domains tested. While distribution matching has better and more consistent performance, we identify one case in which coverage maximization is beneficial - when tasks that receive less trained are more important. Overall, our results show that selective experience replay, when suitable selection algorithms are employed, can prevent catastrophic forgetting. 


\end{abstract}

\section{Introduction} \label{intro}

Lifelong machine learning \cite{Thrun1996} examines learning multiple tasks in sequence, with an emphasis on how previous knowledge of different tasks can be used to improve the training time and learning of current tasks. The hope is that, if sequential learning can be repeated indefinitely, then a system can continue to learn over the course of its lifetime. A continual learning agent is thought to be an important step toward general artificial intelligence.

Recent advances in deep learning 
continue to motivate research into neural networks that can learn multiple tasks. Some approaches include training separate networks for each task \cite{rusu2016progressive,yin2017knowledge} and discovering action hierarchies \cite{tessler2016deep}, but these methods become inefficient when scaling to multiple tasks. Efficiently retaining knowledge for every task over the course of a system's lifetime is a core difficulty in implementing a such a network.
When learning multiple tasks in sequence, each new task changes the distribution of experiences, the optima move, and a large set of states are no longer visited. 
As a result of the non-stationary training distribution, 
the network loses its ability to perform well in previous tasks in what has been termed \emph{catastrophic forgetting}  \cite{mccloskey1989catastrophic,ratcliff1990connectionist,goodfellow2013empirical}.     

We focus on deep reinforcement learning where the effects of catastrophic forgetting are even more pronounced. In deep reinforcement learning, experiences are typically stored in a first-in-first-out (FIFO) replay buffer. A network may experience forgetting on a \emph{single task} if rarely occurring events fall out of the buffer \cite{lipton2016combating} or if the network converges to behavior that does not consistently visit the state space \cite{de2015importance}. This can be mitigated by having a replay buffer with limitless capacity or, if the task boundaries are known a priori, by maintaining separate buffers for each task. 
However, as the number of tasks grows large, storing all experiences of all tasks becomes prohibitively expensive, and ultimately infeasible for lifelong learning.



In this paper, we develop a process for accumulating experiences online that enables their long term retention given limited memory. We treat preserving prior ability as an active process where replay of prior tasks are incorporated into the current learning process. In order to efficiently store prior experiences, we propose a rank-based method for the online collection and preservation of a limited set of training examples to reduce the effects of forgetting. 
We then explore four different selection strategies 
to identify a good ranking function for selecting the experiences. To validate the selection process, we apply each strategy to an autonomous driving domain where different tasks correspond to navigating different unsigned intersections. 
This domain has recently been analyzed as a multi-task setting that both benefits from transfer and allows for the specification of a large variety of tasks \cite{isele2017transferring}. We additionally show our results generalize with further testing in a grid world navigation task, and lifelong variant of the MNIST classification problem. In this work, we make the following contributions:

\begin{itemize}[noitemsep]
\setlength\itemsep{0mm}
\item We propose a selective experience replay process to augment the FIFO replay buffer to prevent catastrophic forgetting.



\item We present and compare four selection strategies for preserving experiences over the course of a system's lifetime. 

\item We analyze the trade-offs between the two most successful selection strategies: distribution matching and coverage maximization.

\end{itemize}

The rest of the paper is organized as follows. In Section \ref{sec:lifelong}, we describe the problem of lifelong machine learning and describe how the sequential training of multiple tasks introduces bias in the modern implementations of 
reinforcement learning with experience replay in Section \ref{sec:bias}. In Section \ref{sec:er}, we propose a selective process for storing experiences that can remove the bias in the training process, before discussing specific selection strategies in Section \ref{sec:selection}. We describe our testing domain and experimental setup in Section \ref{sec:experiments}, and discuss the results of our experiments in Section \ref{sec:results} before concluding in Section \ref{sec:conclusion}.

\section{Lifelong Machine Learning}\label{sec:lifelong}

Lifelong machine learning \cite{Thrun1996,Ruvolo2013} deals with learning many tasks over the lifetime of a system. Tasks are encountered sequentially and prior knowledge is preserved and used to improve the learning of future tasks. Previous lifelong learning efforts using a single deep neural network have focused on preserving prior knowledge within the network.

Numerous works in shallow networks investigate catastrophic forgetting for classification tasks \cite{french1999catastrophic}. The first paper to address the topic in deep nets proposed a local winner-take-all (LWTA) activation to allow unique network configurations to be learned for different tasks \cite{srivastava2013compete}. The LWTA strategy of using different subsets of the network draws parallels to dropout, which was shown to be superior to LWTA at preventing catastrophic forgetting in some domains \cite{goodfellow2013empirical}. Most recently, the Fisher information matrix of previous tasks has been incorporated into the objective function to selectively slow down learning on important weights \cite{kirkpatrick2016overcoming}.  

We instead consider a complementary method to these techniques by selecting and preserving experiences for future training. The idea of retraining on old experiences has been briefly suggested for a similar purpose in training recurrent neural networks \cite{schmidhuber2015learning}. It has  been shown that experience selection can improve training in the context of selecting samples for training \cite{schaul2015prioritized}, but that did not look at the more permanent decision of selecting which samples to keep and which samples to discard. De Bruin et al. [\citeyear{de2016improved}] showed some evidence to suggest that retraining on old experiences can benefit transfer, but did not investigate the effects of training over repeated transfers. We believe replaying experiences is a necessary component to consolidating knowledge from multiple tasks. By preserving experiences from prior tasks and incorporating them into the learning process, we are able to overcome the bias that exists when training on tasks in sequence. Additionally, our approach is agnostic to task boundaries, putting less burden on the system to identify the changes of a non-stationary distribution.

\subsection{Bias in Lifelong Learning}\label{sec:bias}

We consider a reinforcement learning agent and describe how sequentially training on multiple tasks introduces bias into the learning process. In reinforcement learning (RL), an agent in state $s$ takes an action $a$ according to the policy $\pi_\theta$ parameterized by $\theta$. 
The agent transitions to the state $s'$
, and receives a reward $r$. This collection is defined as an experience  $e = (s,a,r,s')$. 

The goal at any time step $t$ is to maximize the future discounted return $R_t = \sum_{k=t}^T \gamma^{k-t} r_{k}$ with discount factor $\gamma \in (0,1]$. Q learning \cite{watkins1992q} defines an optimal action-value function $Q^*(s,a)$ as the maximum expected return that is achievable when taking action $a$ and then following the optimal policy. If $Q^*(s',a')$ is known for all possible actions $a'$, then $Q^*(s,a)$ can be calculated by selecting the max action $a'$, $Q^*(s,a) = \mathbb{E}[r + \gamma \max_{a'} Q^*(s',a')|s,a]$. 
The Bellman equation then provides an iterative update for learning the Q function, $Q_{i+1}(s,a) = \mathbb{E} [r + \gamma \max_{a'} Q_i(s',a')|s,a]$. This can be modified into the loss for an individual experience in a deep Q-network (DQN) as   

\begin{eqnarray}
\mathcal{L}(e_i,\theta) = \bigg( r_i + \gamma \max_{a_i'}Q(s_i',a_i';\theta) - Q(s_i,a_i;\theta)\bigg)^2    \enspace .
\end{eqnarray}
In a lifelong learning setting, this loss yields the multi-task objective
\begin{eqnarray}
	\frac{1}{m}\sum_{j=1}^m \frac{1}{n_j}\sum_{i=1}^{n_j}\mathcal{L}(e_i,\theta) \enspace ,
\end{eqnarray}
where $n_j$ is the number experiences for task $j$, $m$ is the total number of tasks, and $\theta$ is common across for all tasks.
In multi-task learning, it is typically assumed that each task is drawn from a different distribution. This causes experiences within task to be correlated with respect to the global distribution. 

\begin{figure}[thpb!]
    \centering
    \hspace{-10pt}
    	\includegraphics[trim={0mm 15mm 0mm 15mm}, clip, height=1.25in]{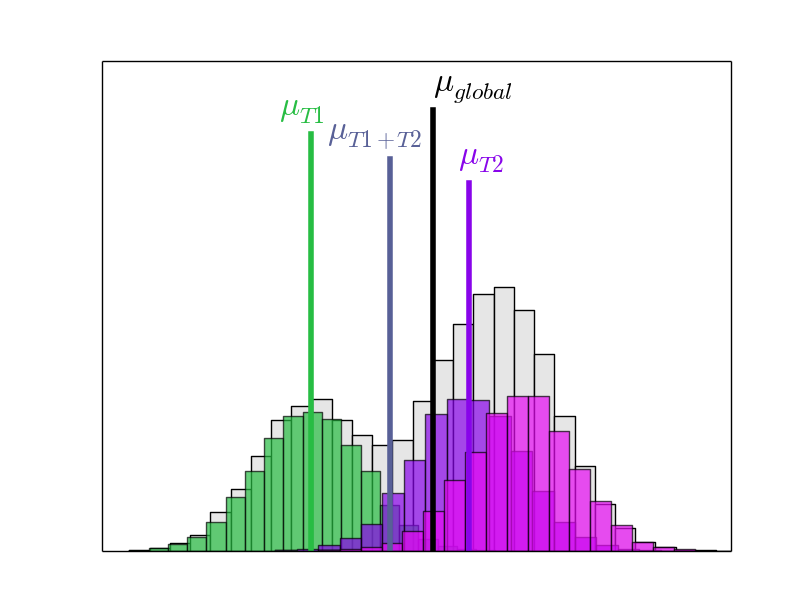}
        \caption{Cartoon of how training on a single task introduces bias. An estimator that only uses data from task 1 produces a biased estimate of the global mean. If the estimator is allowed to accumulate data from more tasks it will approach the true mean, but if it forgets previous tasks the estimate will endlessly track the current task.}\label{fig:bias}
       
\end{figure}
Consider trying to model the global mean given only data from a single task for an estimator $U$,
\begin{eqnarray}
	\text{bias} = \mathbb{E}(U) - \mu_{global} \enspace .
\end{eqnarray}
As more tasks are incorporated into the estimator, the bias will reduce. But if the estimator always forgets the previous tasks, the estimator will track the current task and never converge to the true mean. 





Experience replay in reinforcement learning \cite{lin1992self} has been noted to remove temporal correlations in the data that can bias the estimators towards the current samples \cite{mnih2015human}. 
However, when a system is learning a collection of tasks, the training time on a single task can force experiences from previous tasks out of the buffer, and as a result, the FIFO implementation of experience replay does not address task related correlations. 



When training a single network on a sequence of tasks, we could preserve experiences from previous tasks by maintaining a buffer for each task or by preserving a single buffer with unlimited capacity. Experiences can then be incorporated into the learning process by sampling from previous tasks, and by preserving experiences for every task, the network can optimize all tasks in parallel. 
However, preserving experiences 
for all tasks will quickly run into memory constraints.  
Assuming limited capacity, the standard FIFO implementation of an experience replay buffer will lose earlier experiences and the training distribution will drift over time.  


In order to address this problem, we propose a selective process for storing experiences. Through the use of an experience replay storage mechanism that selectively preserves experiences in long term memory, we can eliminate the non-stationary effects caused by using a FIFO replay buffer.

\section{Selective Experience Replay}\label{sec:er}

In biological systems, experience replay is an important aspect of learning behaviors. Research shows that blocking replay disrupts learning \cite{girardeau2009selective,ego2010disruption} and some suggest that replay is a necessary component of memory consolidation \cite{carr2011hippocampal}. 

The idea that experience replay could be beneficial in machine learning was first explored by Lin [\citeyear{lin1992self}] who used replay in order to speed up training by repeating rare experiences. More recently, experience replay has seen widespread adoption, since it was shown to be instrumental 
in the breakthrough success of deep reinforcement learning \cite{mnih2013playing}. But in its current form, experience replay buffers are typically implemented as randomly sampled first-in-first-out (FIFO) buffers \cite{mnih2015human,van2016deep,lillicrap2015continuous}. 

\subsection{Experience Selection}\label{sec:selection}

In order to efficiently store a sample population that preserves performance across tasks, we introduce a long term storage, \emph{episodic memory}, which accumulates experiences online, storing a subset of the system's history under the restriction of resource constraints. While episodic memory could involve generative processes \cite{goodfellow2014generative} or compression \cite{argyriou2008convex}, in this work we focus specifically on the case of selecting complete experiences 
for storage. 

Episodic memory accumulates knowledge over time which helps to preserve past performance and constrains the network to find an optima that respects both the past and present tasks. To accumulate a collection of experiences from all tasks we rank experiences according to importance for preserving prior knowledge. As long as the ranking function is independent of time, the stored experiences will be as well. In practice we implement episodic memory with a priority queue, ensuring that the least preferred experiences are the first to be removed. Episodic memory serves as long term storage, and is not intended to replace the short term buffer. 

\subsection{Keeping the FIFO Buffer}

In addition to the episodic memory, we keep a small FIFO buffer that constantly refreshes to ensure that the network has the opportunity to train on all incoming experiences. This separation relates to work on complementary learning systems \cite{mcclelland1995there}.
A system with limited storage that preserves memories over an entire lifetime will have very little capacity for new experiences and using long term only storage will impair the learning of new tasks\footnote{We experimented training the network using long term storage only and confirmed this hypothesis. The network performs acceptably, but does not match the performance of the combined approach. See details in the appendix.}. Additionally, by increasing the variety of experiences to which the network is exposed we prevent the network from over-fitting. Note that only the long term memory will hold experiences for all tasks. By training the network on both the long term and short term memories we bias the gradients towards prior tasks. We now turn our attention to the problem of identifying an appropriate ranking function for selecting which experiences to preserve in long term memory.

\subsection{Selection Strategies}

There are several reasonable solutions when faced with the design choice of which experiences to store in the episodic memory. Neuroscience literature suggests that replay occurs most often when events are infrequent, and the breadth of replayed experiences are not limited to the most recent events experienced nor the current task being undertaken \cite{gupta2010hippocampal}. Notably replayed events show preference for 
surprising \cite{cheng2008new} and rewarding situations \cite{atherton2015memory}.  
Alternatively, we can use statistical strategies. We can seek to select experiences that match 
the distribution of all tasks \cite{bickel2009transfer}, or, given that the future test distribution may differ from the training distribution, we can look to optimize coverage of the state space. 
For the rest of this section, we examine selection strategies based on surprise, reward, distribution matching, and coverage - comparing them to the traditional FIFO memory buffer. Each strategy gives a different ranking function $\mathcal{R}(\cdot)$ to order experiences.  

\paragraph{Surprise}

Unexpected events have been shown to be connected to replay in rodents \cite{mcnamara2014dopaminergic,cheng2008new} and \emph{surprise} has a convenient counterpart in incremental reinforcement learning methods like Q-learning. 
The temporal difference (TD) error describes the prediction error made by the network and can be interpreted as the surprise of a particular transition 
\begin{eqnarray}
\mathcal{R}(e_i) = |r_i + \gamma \max_{a'}Q(s_i',a') - Q(s_{i},a_{i}) | \enspace.
\end{eqnarray}
This idea is similar to the idea of uncertainty sampling in active learning \cite{settles2010active}, identifying unexpected inputs in history compression \cite{schmidhuber1993learning}, and it has been effectively used as the foundation for a prioritized sampling strategy \cite{schaul2015prioritized}. 

To store surprising samples, we use a priority queue ordered by the absolute value of the TD error. While the distribution of large errors are not necessarily distributed uniformly throughout the learning process, in practice our experiments show large errors occur throughout training - see Appendix.    

\paragraph{Reward}
Neuroscience research suggests replay occurs more often for rewarding events \cite{atherton2015memory,olafsdottir2015hippocampal}. Prioritized sampling that favors rewarding transitions has also been shown to be helpful in RL \cite{jaderberg2016reinforcement}. We implement this by using the absolute value of the future discounted return of a given experience
\begin{eqnarray}
\mathcal{R}(e_i) = |R_i(e_i)| \enspace.
\end{eqnarray}

\paragraph{Global Distribution Matching}

We can expect the best test performance if the training distribution matches the test distribution. This motivates a strategy that tries to capture the combined distribution of \emph{all} tasks. To turn this into a ranking function, we use random selection to down sample the set of experiences to a smaller population that approximately matches the global distribution. 
In order to maintain a random sample over the global distribution even though experiences arrive sequentially and the global distribution is not known in advance, we use a reservoir sampling strategy that assigns a random value to each experience. This random value is used as the key in a priority queue where we preserve the experiences with the highest key values. 
\begin{eqnarray}
\mathcal{R}(e_i) \sim N(0,1)
\end{eqnarray}
At time $t$ the probability of any experience being the max experience is $1/t$ regardless of when the sample was added, guaranteeing that at any given time the sampled experiences will approximately match the distribution of all samples seen so far
\begin{eqnarray}
\forall e_i ~ p(e_i\in B) = \mbox{min}\left(1,\frac{|B|}{t}\right) \enspace .
\end{eqnarray}
Here $|B|$ is the number of experiences stored in episodic memory $B$. 

\paragraph{Coverage Maximization} 
When limiting the number of preserved experiences, it might also be advantageous to maximize coverage of the state space. Research shows that limited FIFO buffers can converge to homogeneous transitions when the system starts consistently succeeding, and that this lack of error samples can cause divergent behavior \cite{de2015importance}. While it is often not possible to know the complete state space, de Bruin et al. [\citeyear{de2016improved}] suggest that approximating a uniform distribution over the visited states can help learning when buffer size is limited. 


In order to maximally cover the space, we approximate a uniform distribution by ranking experiences according to the number of neighbors in episodic memory within a fixed distance $d$ and replacing entries with the most neighbors. 
This approach makes the time complexity of adding entries 
$O(log(|E|))$ where $|E|$ is the number of stored experiences in episodic memory. This time complexity can be reduced to $O(log(|E|))$ by using a KD-tree \cite{bentley1975multidimensional} and only comparing against the $k$ nearest neighbors or to $O(k)$ by sampling the buffer.    
\begin{align}
& \mathcal{N}_i = \{e_j ~s.t.~ \mbox{dist}(e_i - e_j) < d \} \\
& \mathcal{R}(e_i) = -| \mathcal{N}_i |, \mbox{order according to rank} 
\end{align}
To confirm that this method does indeed give us good coverage, we project the state space down to two dimensions using the non-linear randomized projection t-SNE \cite{maaten2008visualizing}. This projection also visualizes how different tasks come from different distributions in the state space. Since the selection process is done in the high dimensional space, not the mapped space, we do not expect the selected points to look perfectly uniform after projection, however we can show that the algorithm does indeed cover the space, which we show in the appendix. 

\section{Experiments} \label{sec:experiments}

\begin{figure*}[t!]
    \centering
    \hspace{-14pt}
    \begin{subfigure}[b]{1.1in}
    	\includegraphics[height=.7\textwidth]{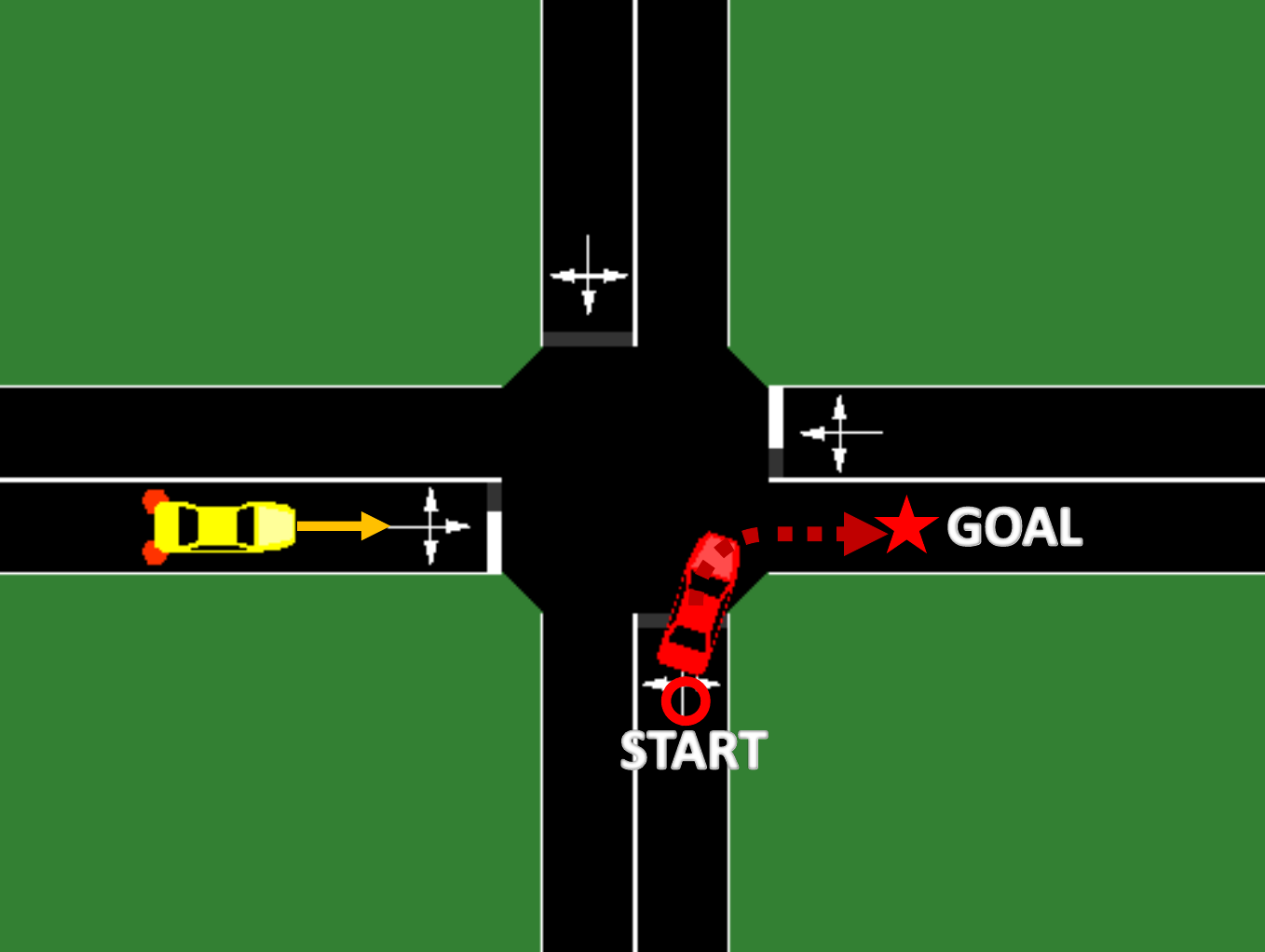} 
        \caption{\emph{Right}}
        \label{fig:scenarios_right}
    \end{subfigure}
    \begin{subfigure}[b]{1.1in}
    	\includegraphics[height=.7\textwidth]{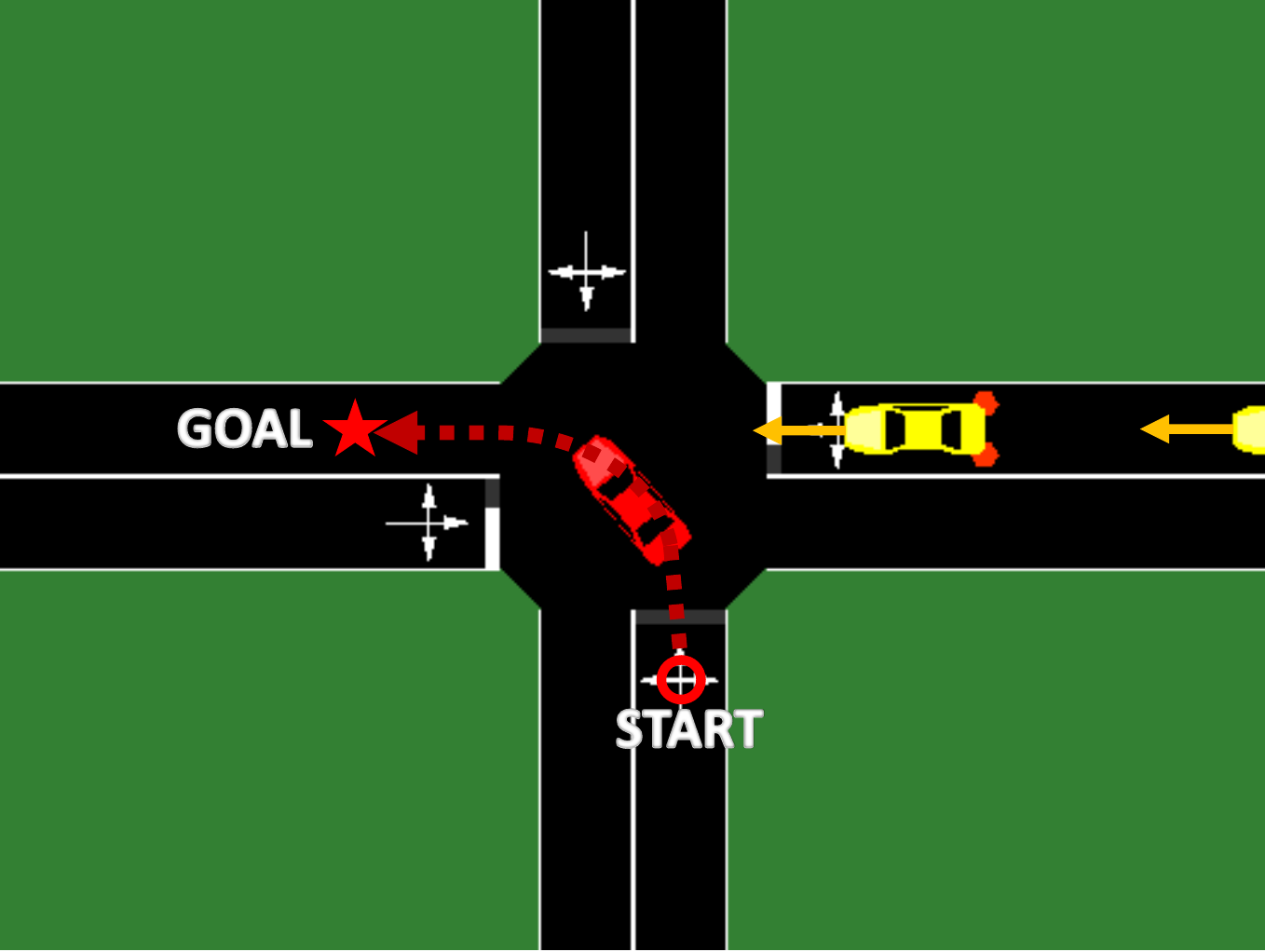}
        \caption{\emph{Left}}
        \label{fig:scenarios_left}
    \end{subfigure}
    \begin{subfigure}[b]{1.1in}
    	\includegraphics[height=.7\textwidth]{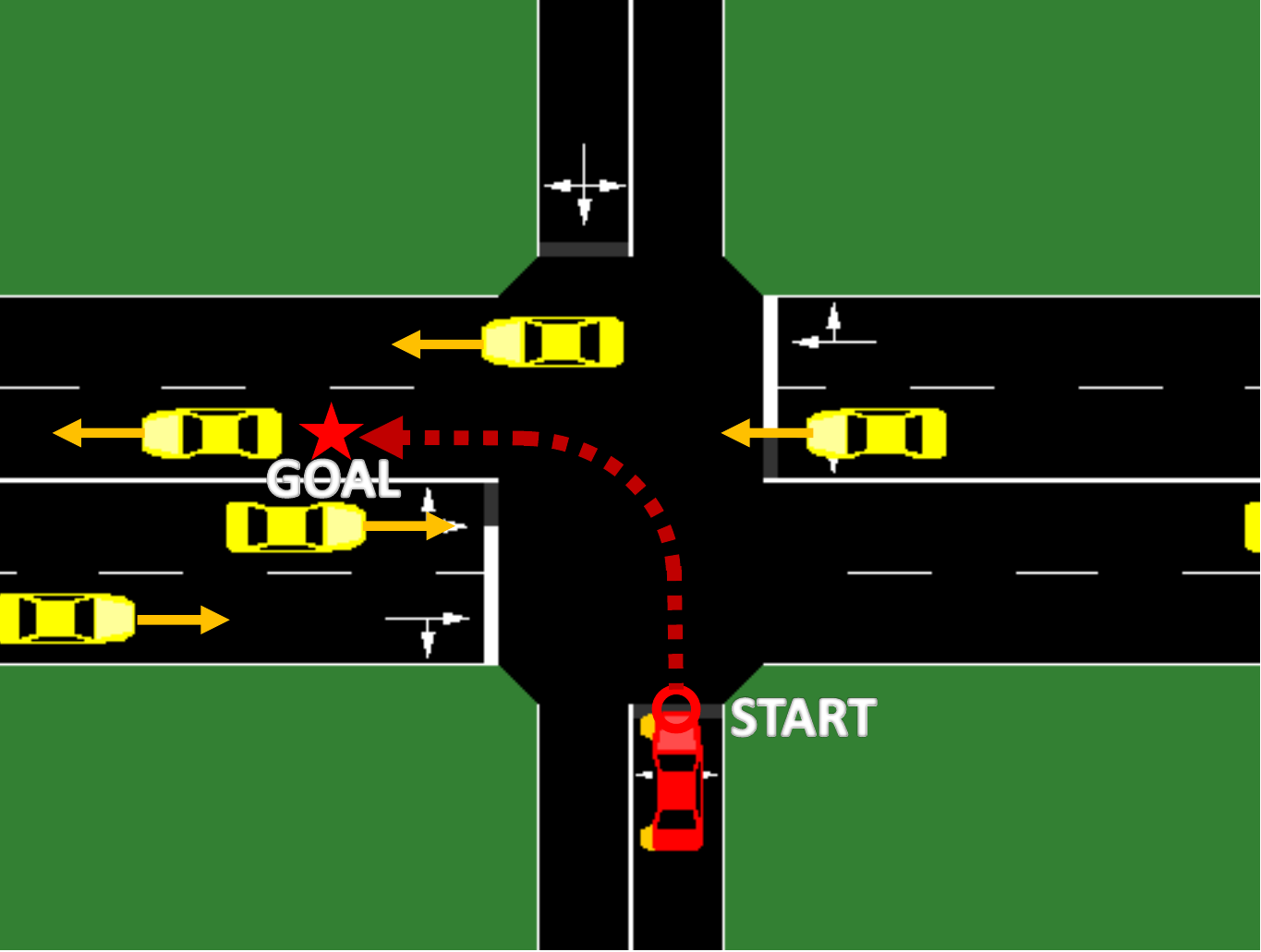}
        \caption{\emph{Left2}}
        \label{fig:scenarios_left2}
    \end{subfigure}
    \begin{subfigure}[b]{1.1in} 
       \includegraphics[height=.7\textwidth]{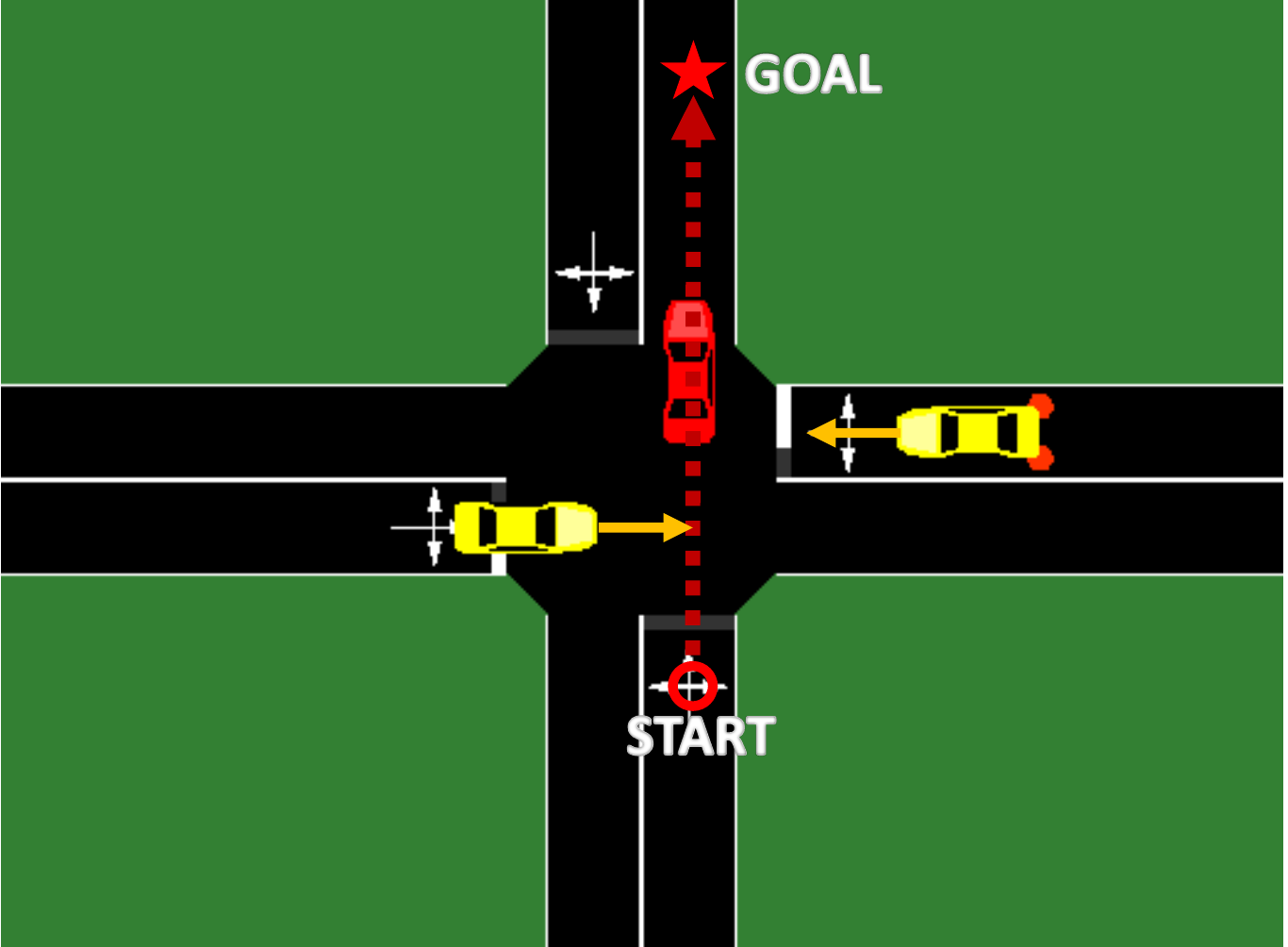}
        \caption{\emph{Forward}}
        \label{fig:scenarios_forward}
    \end{subfigure}
    \begin{subfigure}[b]{1.1in}
    	\includegraphics[height=.7\textwidth]{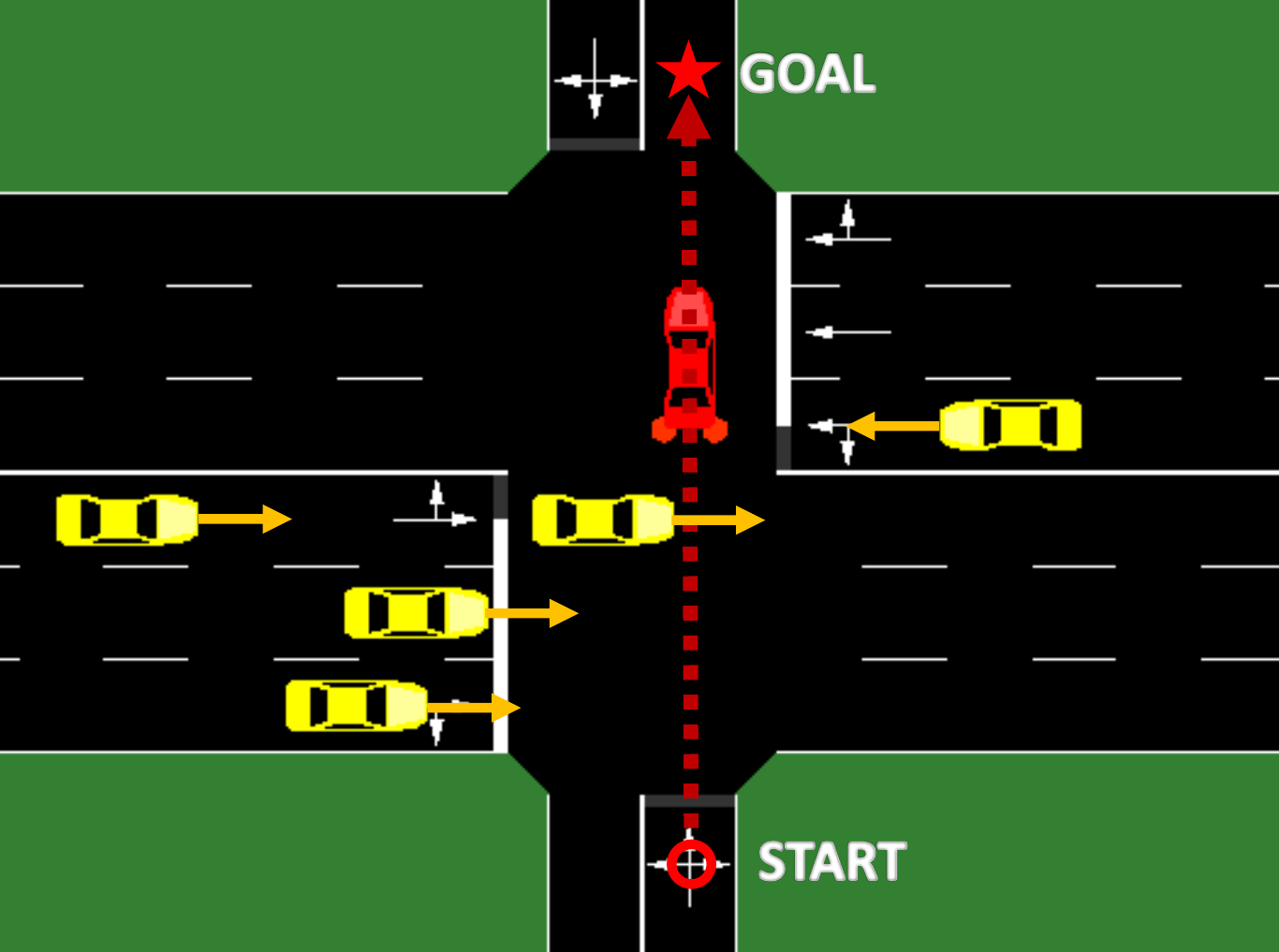}
        \caption{\emph{Challenge}}
        \label{fig:scenarios_challenge}
    \end{subfigure}
        \caption{Visualizations of intersection tasks used for our experiments. }\label{fig:scenarios} 
\end{figure*}

We evaluate selective experience replay on the problem of autonomously handling unsigned intersections. 
The Sumo simulator \cite{sumo} allows for the creation of a variety of different driving tasks with varying levels of difficulty, many of which benefit from transfer while being sufficiently different that adaptation is necessary for good performance \cite{isele2017transferring}. 
Autonomous driving has multiple objectives which often conflict such as maximizing safety, minimizing the time to reach the goal, and minimizing the disruption of traffic. The autonomous agent chooses between two types of actions: to go or to wait for some number of iterations. The path to take is assumed to be provided to the agent by a high-level controller. For our experiments we focus on the number of successfully completed trials. Successful trials consist of the car navigating safely through the intersection, considering both collisions and timeouts to be failures. 

We ran experiments using five different intersection tasks: \emph{Right}, \emph{Left}, \emph{Left2}, \emph{Forward} and a \emph{Challenge}. Each of these tasks is depicted in Figure \ref{fig:scenarios}.
The \emph{Right} task involves making a right turn, the \emph{Forward} task involves crossing the intersection, the \emph{Left} task involves making a left turn, the \emph{Left2} task involves making a left turn across two lanes, and the \emph{Challenge} task involves crossing a six lane intersection. 


The order in which tasks are encountered does impact learning, and several groups have investigated the effects of ordering \cite{Bengio2009,ruvolo2013active,narvekar2016source}. For our experiments, we selected a random task ordering that demonstrates forgetting and held it fixed for all experiments. We describe the details of our network and training parameters in the appendix. All of our experiments use the dropout training method \cite{hinton2012improving} which was shown to help prevent forgetting \cite{goodfellow2013empirical}. 

In order to evaluate the effectiveness of the proposed approach, we ran three sets of experiments. In the first experiment, we evaluate the upper bound: a network with an unlimited capacity FIFO buffer, and the lower bound: a limited memory FIFO buffer. We sequentially train a network on five intersection tasks with a fixed order, $10,000$ training experiences on each task. We evaluate and plot the success rate at periodic intervals during training. In the second experiment, we investigate the performance of selective experience replay with a limited size experience replay buffer. We train a different network for each of the four selection strategies and compare the performance and check whether the network undergoes catastrophic forgetting during training. In the third experiment, we compare the two most promising strategies when the number of training samples are not evenly distributed across tasks. For each strategy, we train a network on the \emph{right} task for $2000$ experiences followed by $25,000$ experiences on the $challenge$ task. These tasks were selected as being the most different from each other based on previous experiments. We compare the final network on task performance for evaluation. The next section presents the results of our experiments.

To confirm that our results generalize to other domains and beyond reinforcement learning we additionally test our results on a grid world navigation task an a lifelong learning variant of MNIST \cite{lecun1998gradient}. 

In the grid world domain the world is divided into four rooms. The agent starts in one room and navigates to a random goal position in one of the other rooms. Each tasks corresponds to the room which contains the goal. In the lifelong learning variant of MNIST, the agent is exposed to only two digits. At the end of training on five tasks the agent is expected to be able to correctly classify all 10 digits. Further details on both experiments can be found in the appendix.

\section{Results}\label{sec:results}

\subsection{Baselines}

\begin{figure}[thpb!]
    \centering
    \hspace{-10pt}
    \begin{subfigure}[b]{1.6in}    	\includegraphics[trim={4mm 0 0 13mm}, clip, height=1.2in]{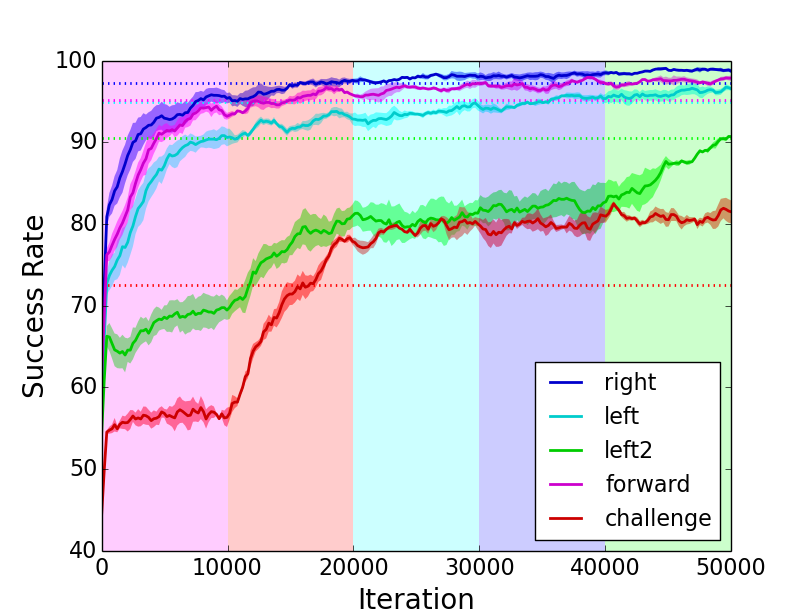}
        \caption{Unlimited capacity}
        \label{fig:LL}
    \end{subfigure}
    \begin{subfigure}[b]{1.6in}
    	\includegraphics[trim={14mm 0 0 13mm}, clip, height=1.2in]{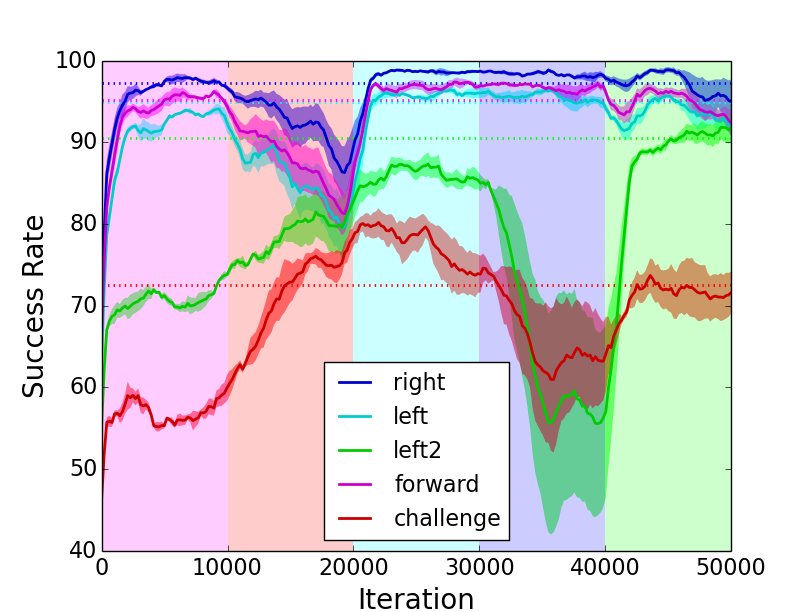}
        \caption{Limited capacity (FIFO)}
        \label{fig:FIFO}
    \end{subfigure}
        \caption{(a) Lifelong learning of multiple tasks when the full history of experiences is available for each task. The dotted lines indicate the final performance of the single task network with an equivalent amount of training. (b) Training with a FIFO experience replay buffer on multiple tasks with limited capacity. The traditional method of storing memories for experience replay demonstrates catastrophic forgetting.}\label{fig:baselines}
        \vspace{-10pt}
\end{figure}

Figure \ref{fig:LL} shows the performance of a network training with an unlimited capacity replay buffer.
There are five tasks in this example, horizontal dotted lines indicate the final performance of a single task network trained for an equivalent number of training steps. The learning curves indicate the performance of a single lifelong learning network training on all tasks. Each color indicates the performance on a different task. The background color indicates the task the network is currently training on. Error bars show the standard deviation across three trials where a test at each time step involves 100 trials. 
We see that when the experience buffer has unlimited capacity, the network improves over the performance of the single task networks given an equivalent amount of training.
Training on a new task receives the \emph{jumpstart} \cite{taylor2009transfer} benefit of prior knowledge and the network is resistant to forgetting. However, training with unlimited capacity is not practical as the number of tasks increases. This experiment demonstrates the upper bound performance we are hoping to match when the experience replay buffer capacity is limited.

When a FIFO replay buffer has limited capacity (1000 experiences), previous tasks are forgotten (Figure \ref{fig:FIFO}). We observe that training on the \emph{forward} task initially boosts every tasks performance, however when the network begins training on the challenge task, performance drops on all single lane tasks (\emph{right}, \emph{left}, and \emph{forward}). Similarly, training on the \emph{right} task hurts the performance of other tasks. 

\begin{figure*}[tb!]
    \centering
    \hspace*{-10pt}
    \begin{subfigure}[b]{1.6in}
    	\hspace*{-10pt}
    	\includegraphics[trim={4mm 0 0 13mm}, clip, height=1.25in]{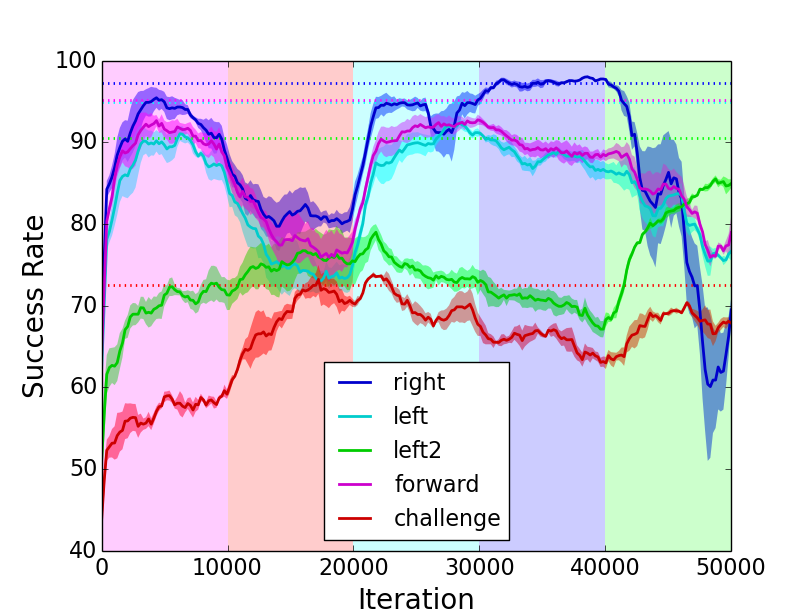}
        \caption{Surprise}
        \label{fig:TD}
    \end{subfigure}
    \begin{subfigure}[b]{1.6in}
    	\includegraphics[trim={16mm 0 0 13mm}, clip, height=1.25in]{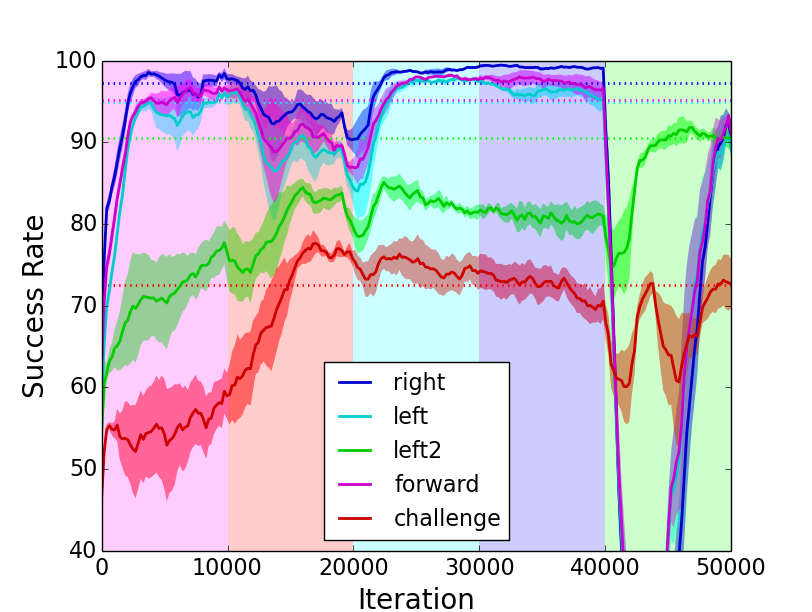}
        \caption{Reward}
        \label{fig:reward}
    \end{subfigure}
    \begin{subfigure}[b]{1.6in}
    	\includegraphics[trim={16mm 0 0 13mm}, clip, height=1.25in]{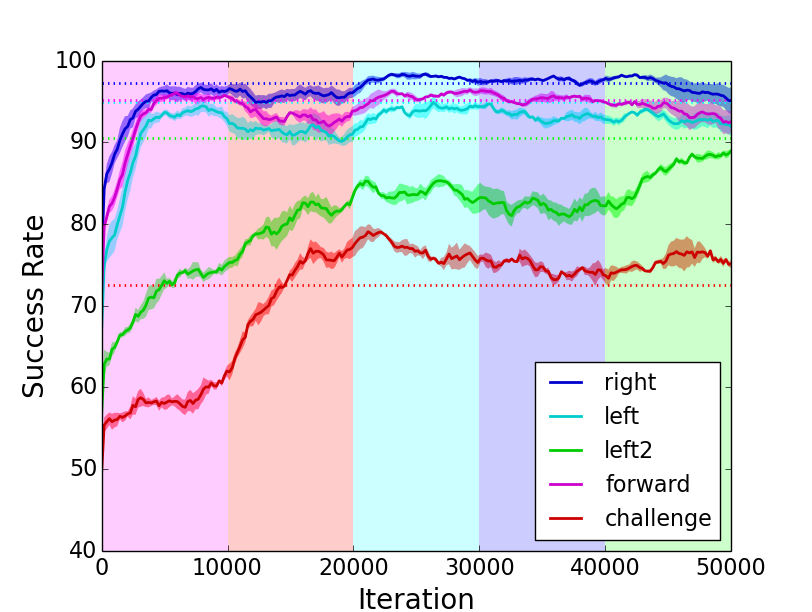}
        \caption{Coverage maximization}
        \label{fig:coverage}
    \end{subfigure}~
     \begin{subfigure}[b]{1.6in}
    	\includegraphics[trim={16mm 0 0 13mm}, clip, height=1.25in]{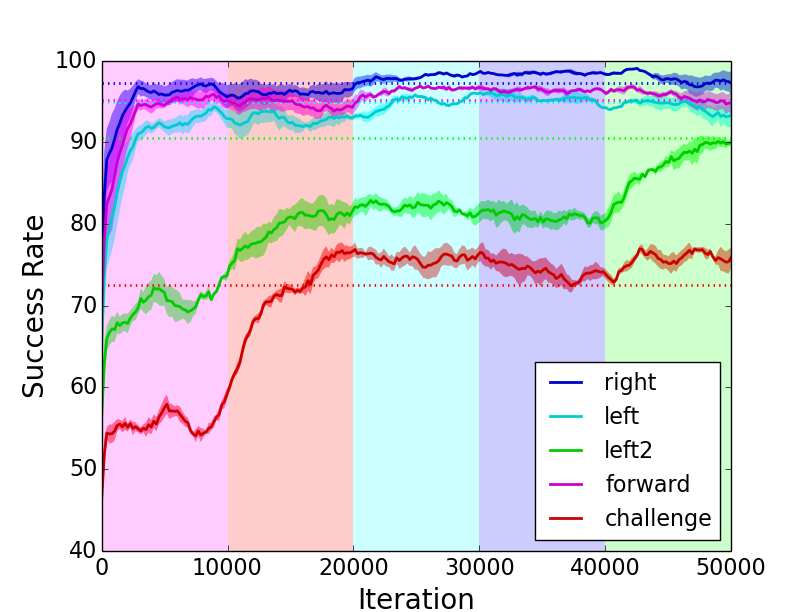}
        \caption{Distribution matching}
        \label{fig:random}
    \end{subfigure}
    \caption{Selective experience replay results. While surprise and reward selection strategies exhibit catastrophic forgetting on previously learned tasks, coverage maximization and distribution matching strategies do not.}\label{fig:selective}
\end{figure*}

\subsection{Selective Replay Results} 

The results of the networks using selective replay are shown in Figure \ref{fig:selective}. 
Our results show that both the surprise and reward strategies behaved poorly as experience selection strategies as they exhibit catastrophic forgetting. We believe that surprise performed poorly as a selection method, because surprising experiences are exactly the samples the system is most uncertain about. It makes sense to focus on these examples to improve learning, however when it comes to retaining knowledge, surprise represents the information the system has \emph{not} learned. When the goal is to preserve memory, experiences where the network makes errors might be very poor experiences to store\footnote{We also ran exploratory tests where we preserved experiences with the minimal TD error and the results were not favorable.}. In the reward selection method, only rewarding experiences are stored. Training on only rewarding experiences ignores the credit assignment problem. Added noise was kept small, and only intended to break ties. Increasing the noise will select states around the rewarding experiences as well and will become a combination of the reward and distribution matching strategies. 

Both the distribution matching and coverage maximization strategies prevented catastrophic forgetting, but they display somewhat different behaviors. By maximizing coverage, frequently visited regions in the state space can be covered with only a few samples, so maximizing coverage will be under-represented compared to distribution matching in places where the system spends the most time. Similarly, strategies that seek to maximize coverage will favor outliers and rare events. This can be helpful if these error cases are failure states that would be otherwise missed \cite{lipton2016combating}. On physical systems, exhibiting a large amount of exploration can quickly wear and damage components, so preserving these experiences for replay can also increase the safety of learning \cite{de2015importance}. In our domain, distribution matching exhibited slightly more stable behavior and had a slight advantage in performance. However, given the differences of the two approaches, it is likely different domains could change the better performing strategy. 


\subsection{Unbalanced Training}
\begin{figure}[bhp!]
\centering
    \hspace{-5pt}
    \begin{subfigure}[b]{1.6in}
    	\includegraphics[trim={8mm 0 0 13mm}, clip, height=1.25in]{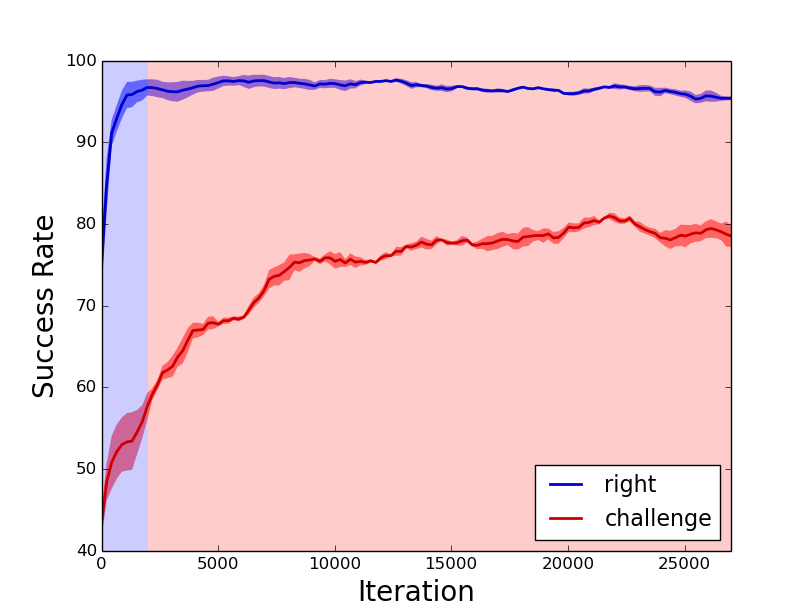}
        \caption{Coverage Maximization}
        \label{fig:imbal_uniform}
    \end{subfigure}
    \begin{subfigure}[b]{1.6in}
    	\hspace*{5pt}
    	\includegraphics[trim={16mm 0 0 13mm}, clip, height=1.25in]{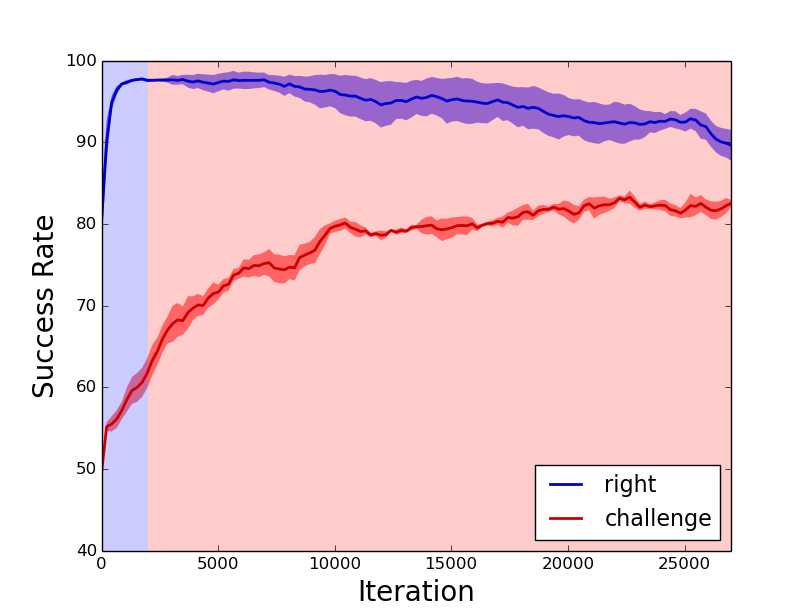}
        \caption{Distribution Matching}
        \label{fig:imbal_rand}
    \end{subfigure}
        \caption{Unbalanced training. The network is trained for 2000 iterations on \emph{right} and then trained for 25000 iterations on \emph{challenge}. Background color indicates the training task.}\label{fig:imbalanced} 
\end{figure}
While the distribution matching strategy slightly outperformed  coverage maximization in the previous experiment, that setup assumes that each task receives equal training time. It is reasonable to hypothesize that this strategy will under perform if we break the assumption that the importance of the task is proportional to the time spent training.

In this experiment, we consider unbalanced training time. We train a single network like previous experiments, but with only two sequential tasks rather than five in order to test the hypothesis. Both networks were trained on \emph{right} task first, then on \emph{challenge} task for 12.5 times longer.

The results of the unbalanced training is shown in Figure \ref{fig:imbalanced}. On the \emph{challenge} task, both strategies performed about the same. On the \emph{right} task, coverage maximization retained better performance. The reason was that the coverage maximization strategy kept more experiences from the \emph{right} task in episodic memory. It could be the case that certain experiences, while only encountered briefly, are valued more. In this case we would expect the strategy that covers the state space more to perform better. We observed that distribution matching strategy had higher variance on the \emph{right}, we think this is due to randomness in selection.


\subsection{Grid World}
\begin{figure*}[tb!]
    \centering
    \hspace{-10pt}
    \begin{subfigure}[b]{1.35in}
    	\includegraphics[trim={0 0 0 10mm}, clip, height=1.0in]{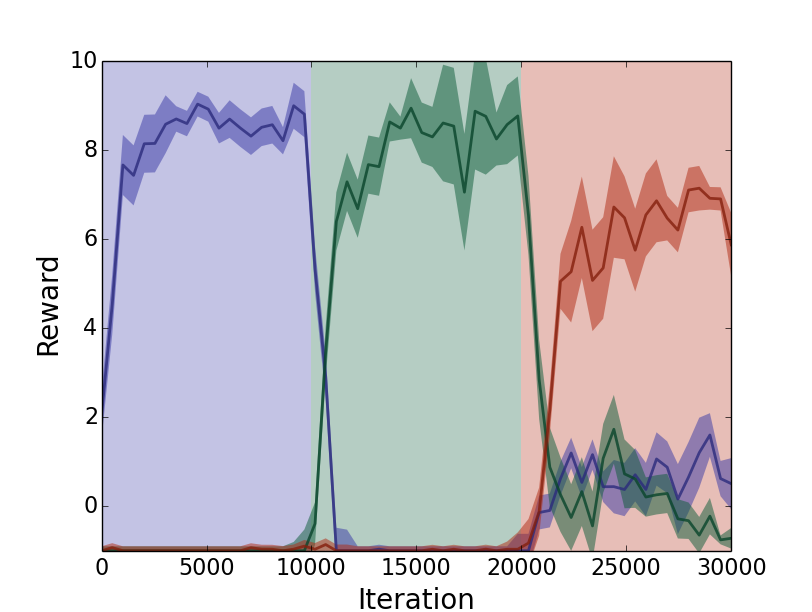}
        \caption{No Selection}
        \label{fig:TDg}
    \end{subfigure}
    \begin{subfigure}[b]{1.35in}
    	\includegraphics[trim={0 0 0 10mm}, clip, height=1.0in]{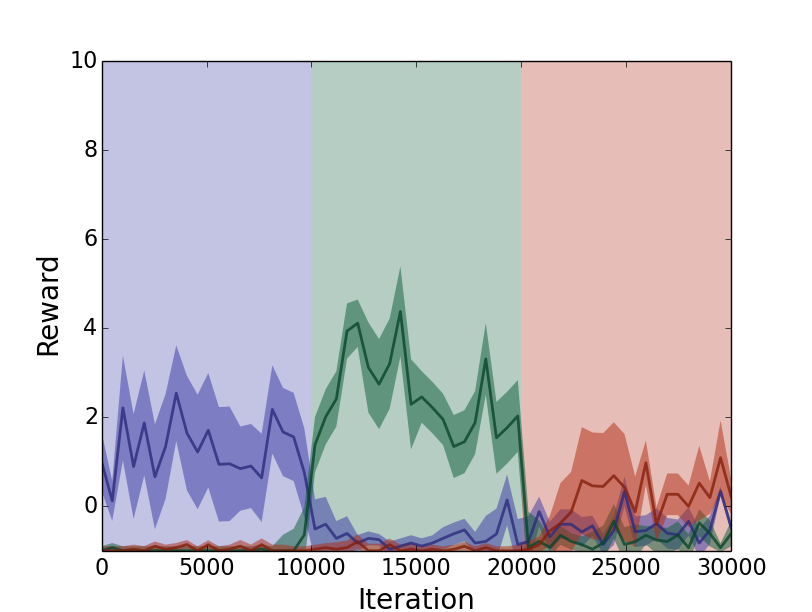}
        \caption{Suprise}
        \label{fig:surpriseg}
    \end{subfigure}
        \begin{subfigure}[b]{1.35in}
    	\includegraphics[trim={0 0 0 10mm}, clip, height=1.0in]{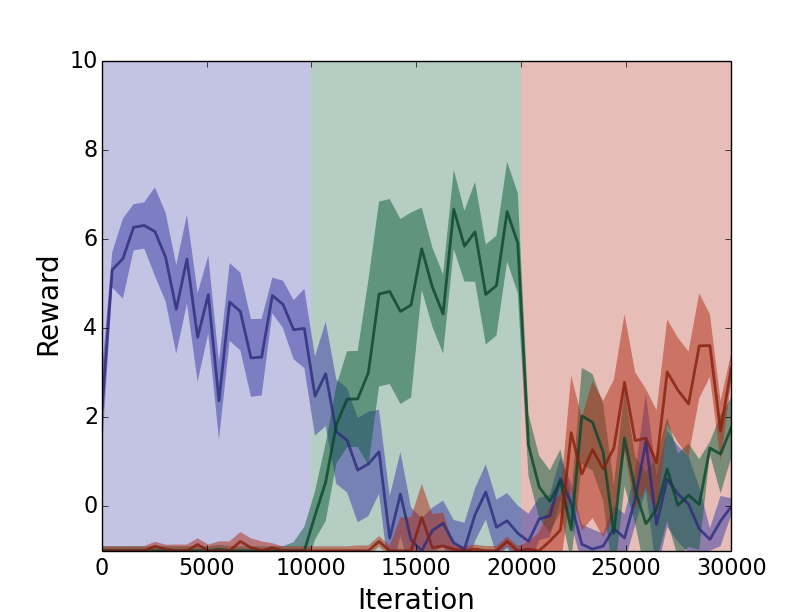}
        \caption{Reward}
        \label{fig:rewardg}
    \end{subfigure}
    \begin{subfigure}[b]{1.35in}
    	\includegraphics[trim={0 0 0 10mm}, clip, height=1.0in]{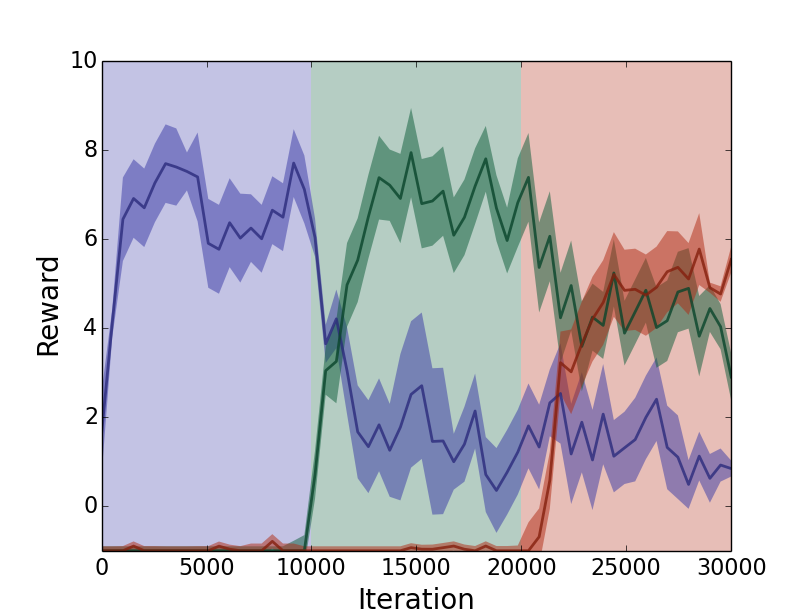}
        \caption{Coverage Max.}
        \label{fig:coverageg}
    \end{subfigure}~
     \begin{subfigure}[b]{1.35in}
    	\includegraphics[trim={0 0 0 10mm}, clip, height=1.0in]{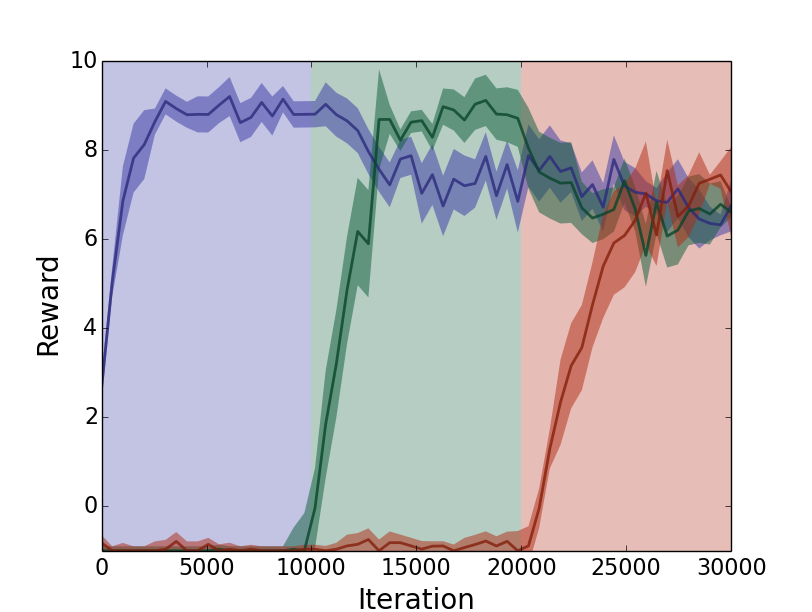}
        \caption{Distribution Matching}
        \label{fig:randomg}
    \end{subfigure}
    \caption{Selective experience replay in the grid world domain. Without selection, performance on previous tasks almost immediately drops to zero. While surprise and coverage maximization help somewhat, distribution matching is the best at preserving experience.}\label{fig:grid}
\end{figure*}

Grid World shows much more dramatic catastrophic forgetting, where the agent almost immediately forgets performance of previous tasks without selection. Results are shown in Figure \ref{fig:grid}. Since the final task passes through the two earlier rooms we note some recovery, but this is probably due to the configuration of the rooms. Surprise and Reward again both perform poorly. Coverage maximization preserves some performance on earlier tasks, but distribution matching both reaches higher performance on current tasks and preserves more performance on earlier tasks.  

\begin{figure*}[tb!]
    \centering
    \hspace{-10pt}
    \begin{subfigure}[b]{1.6in}
    	\includegraphics[trim={0 0 0 10mm}, clip, height=1.25in]{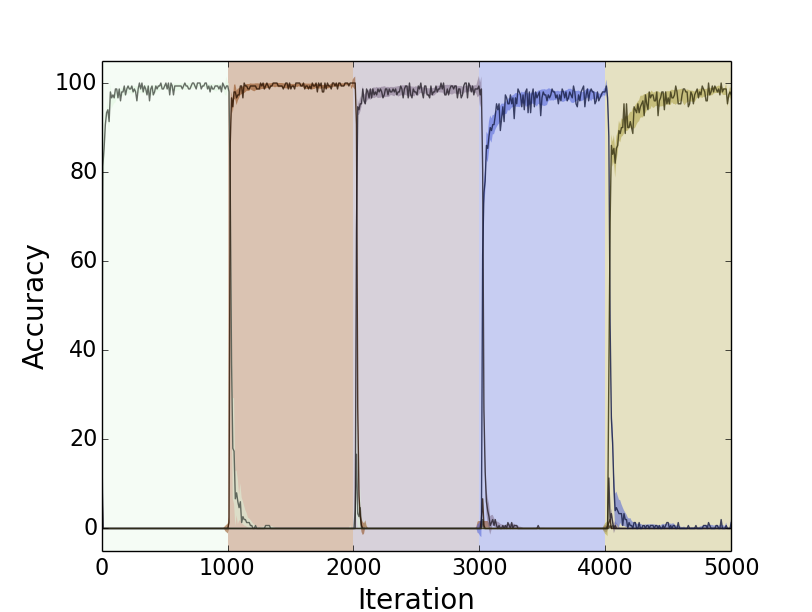}
        \caption{No Selection}
        \label{fig:TDm}
    \end{subfigure}
    \begin{subfigure}[b]{1.6in}
    	\includegraphics[trim={0 0 0 10mm}, clip, height=1.25in]{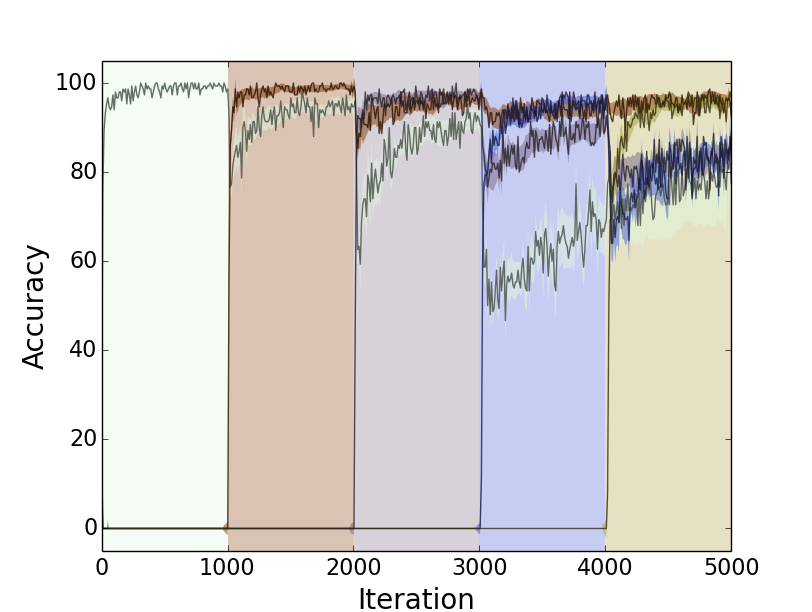}
        \caption{Suprise}
        \label{fig:surprisem}
    \end{subfigure}
    \begin{subfigure}[b]{1.6in}
    	\includegraphics[trim={0 0 0 10mm}, clip, height=1.25in]{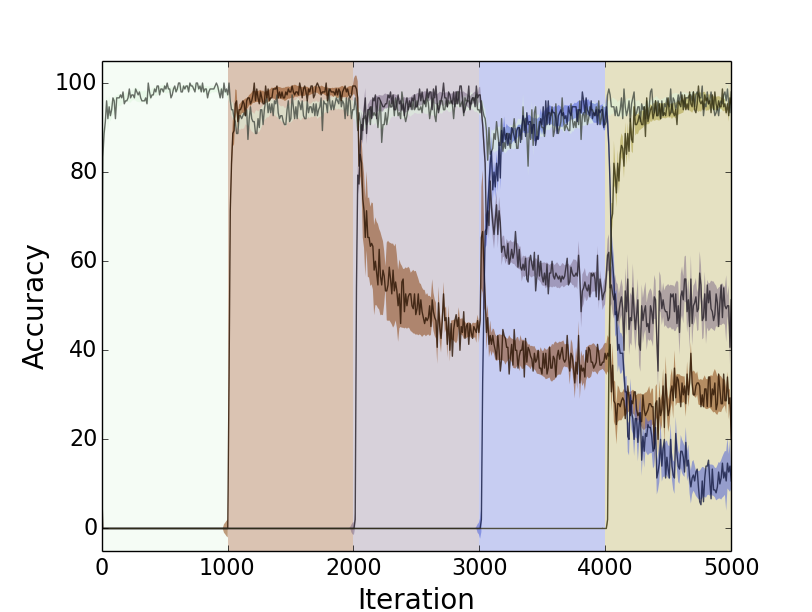}
        \caption{Coverage Maximization}
        \label{fig:coveragem}
    \end{subfigure}~
     \begin{subfigure}[b]{1.6in}
    	\includegraphics[trim={0 0 0 10mm}, clip, height=1.25in]{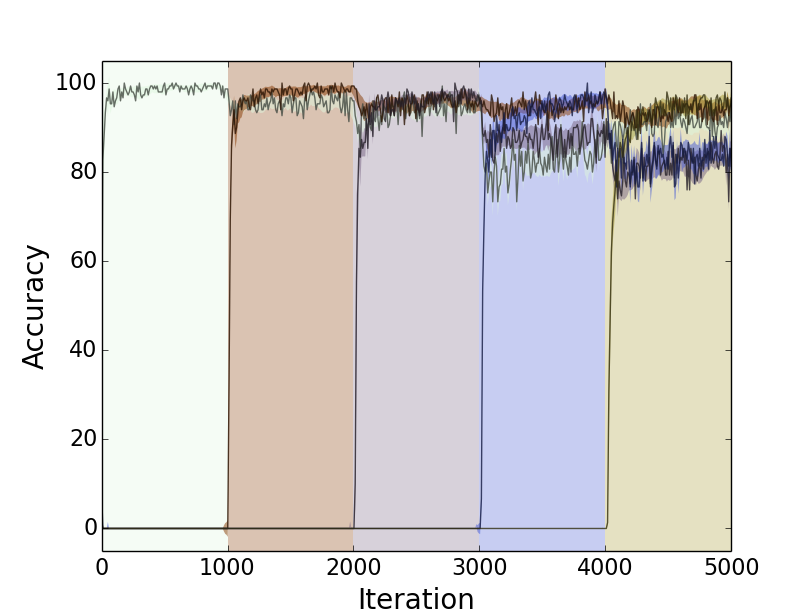}
        \caption{Distribution Matching}
        \label{fig:randomm}
    \end{subfigure}
    \caption{Selective experience replay on MNIST. Without selection, performance on previous tasks almost immediately drops to zero. While surprise and coverage maximization help, distribution matching is the best at preserving experience.}\label{fig:mnist}
\end{figure*}

\subsection{Lifelong MNIST}
The results on the lifelong variant of MNIST are shown in Figure \ref{fig:mnist}. Because the classification space is not shared between tasks, the functions of previous tasks are very quickly destroyed without replay. Here surprise, representing the error in classification, works relatively well at preserving performance. Coverage maximization performs more poorly than usual. This is likely due to difficulties in selecting an appropriate distance metric.

\section{Conclusion}
\label{sec:conclusion}

We proposed a method for selectively retaining experiences to preserve training on previous tasks in a lifelong learning setting. Forgetting previous task knowledge is a major hurdle when applying deep networks to lifelong learning scenarios. Unlimited storage of experiences is intractable for lifelong learning, and the use of FIFO experience buffers has been shown to lead to catastrophic forgetting. Our method involves keeping a long-term experience replay buffer in addition to the short-term FIFO buffer. We compare four strategies for selecting which experiences will be stored in the long-term buffer: a) Surprise, which favors experiences with large TD error, b) Reward, which favors experiences that lead to high absolute reward, c) Distribution Matching, which matches the global training state distribution, d) Coverage Maximization, which attempts to preserve a large coverage of the state space.

We evaluated the four selection methods in an autonomous driving domain where the tasks correspond to handling different types of intersections. In a setting where five intersection tasks were learned sequentially, we found that Surprise and Reward selection methods still displayed catastrophic forgetting, whereas Distribution Matching and Coverage Maximization did not. The latter two strategies achieve nearly comparable performance to a network with unlimited experience replay capacity. Looking at other domains we see that Distribution Matching is consistently the top performer. While distribution matching has better performance, we identify one case in which it is not an ideal strategy - the case where tasks which received limited training are more important. 
Overall, our results show that selective experience replay, when suitable selection algorithms are employed, can prevent catastrophic forgetting in lifelong reinforcement learning.

While the addition of episodic memory is intended to address forgetting in lifelong learning, limited storage and the resulting divergent behavior are also concerns in single task networks and our contribution may find more general application outside of lifelong learning domains. 







\medskip

\begin{small}
\bibliography{refs}
\bibliographystyle{aaai}
\end{small}

\end{document}